\documentclass[sigconf]{acmart}
\usepackage{units}
\usepackage{acronym}
\usepackage[externalcollaborator,year=2022]{jplack}

\acrodef{CO2}[CO$_2$]{carbon dioxide}
\acrodef{H2O}[H$_2$O]{water}
\acrodef{CNN}{convolutional neural network}
\acrodef{HiRISE}{High Resolution Imaging Science Experiment}
\acrodef{PDS}{Planetary Data System}
\acrodef{HEALPix}{Hierarchical Equal Area isoLatitude Pixelation}

\newcommand{\latlon}[2]{$#1^{\circ}\mathrm{N}$,~$#2^{\circ}\mathrm{E}$}

\setcopyright{personal}
\copyrightyear{2022}
\acmYear{2022}
\acmDOI{}

\acmConference[SIGKDD DeepSpatial '22]{3rd ACM SIGKDD Workshop on Deep Learning for Spatiotemporal Data}{August 15, 2022}{Washington, DC}
\acmPrice{}
\acmISBN{}

\begin{document}

\title{Evaluating Terrain-Dependent Performance for Martian Frost Detection in Visible Satellite Observations}

\author{Gary Doran}
\email{gary.b.doran.jr@jpl.nasa.gov}
\orcid{0000-0003-1233-2224}
\author{Serina Diniega}
\orcid{0000-0003-3766-2190}
\author{Steven Lu}
\orcid{0000-0001-6859-7737}
\author{Mark Wronkiewicz}
\orcid{0000-0002-6521-3256}
\author{Kiri L.\ Wagstaff}
\orcid{0000-0003-4401-5506}
\affiliation{%
  \institution{Jet Propulsion Laboratory, \\ California Institute of Technology}
  \streetaddress{4800 Oak Grove Dr}
  \city{Pasadena}
  \state{California}
  \country{USA}
  \postcode{91109}
}

\author{Jacob Widmer}
\orcid{0000-0003-1802-3174}
\affiliation{%
  \institution{University of California, Los Angeles}
  \city{Los Angeles}
  \state{California}
  \country{USA}
}

\renewcommand{\shortauthors}{Doran, et al.}

\begin{abstract}
  Seasonal frosting and defrosting on the surface of Mars is hypothesized to
drive both climate processes and the formation and evolution of geomorphological
features such as gullies. Past studies have focused on manually analyzing the
behavior of the frost cycle in the northern mid-latitude region of Mars using
high-resolution visible observations from orbit. Extending these studies
globally requires automating the detection of frost using data science
techniques such as convolutional neural networks. However, visible indications
of frost presence can vary significantly depending on the geologic context on
which the frost is superimposed. In this study, we (1) present a novel approach
for spatially partitioning data to reduce biases in model performance
estimation, (2) illustrate how geologic context affects automated frost
detection, and (3) propose mitigations to observed biases in automated frost detection.
\end{abstract}

\keywords{planetary science, remote sensing, deep learning}


\maketitle

\section{Introduction}
As on Earth, frost will accumulate on the Martian surface from the poles towards the equator each winter. This frost is an important driver for surface geological and climate processes~\cite{diniega2021geomorph} and provides a key observable constraint for studies of how volatiles are transported around Mars in the present climate~\cite{diniega2020high}. Unlike the Earth, the atmosphere of Mars is comprised primarily of \ac{CO2} and this volatile constitutes most of the frost, falling as snow or condensing at the surface due to surface temperatures falling to the \ac{CO2} frost point. A small amount of water frost will also form when temperatures are below the \ac{H2O} frost point, but only if the local concentration of \ac{H2O} vapor in the atmosphere is high enough. A global, high-resolution map of where and when specific types of frost form around Mars would be a great aid towards generation of a comprehensive view of the Martian global frost cycle and an important input for many studies focused on understanding Mars' atmospheric dynamics, volatile budget, landscape and landform evolution, and surface operations of robotic and human explorers.

Confident detection of Martian frost and characterization of its type (i.e., \ac{H2O} or \ac{CO2}, snowfall or surface condensate) requires the combination of information across multiple remote sensing instruments, including high-resolution visible imaging systems. Over \unit[100]{TB} of high resolution surface image data has been returned from Mars, making it infeasible to manually analyze these images for the presence of frost. Therefore, we have trained a \ac{CNN} model to detect frost using a corpus of labeled data from previously studied frost sites. The model can be deployed across the entire image dataset to automatically detect frost and enable global-scale scientific analysis of the frost cycle.

This paper describes our initial efforts to train and validate a Martian frost detection model for visible images. We describe some domain-specific challenges and approaches related to label collection, validation, and bias characterization. We find that detection recall is biased against certain underrepresented terrain types such as dunes, and we propose future work to mitigate this bias.

\section{Background}

\begin{figure*}
    \centering
    \includegraphics[width=0.19\textwidth]{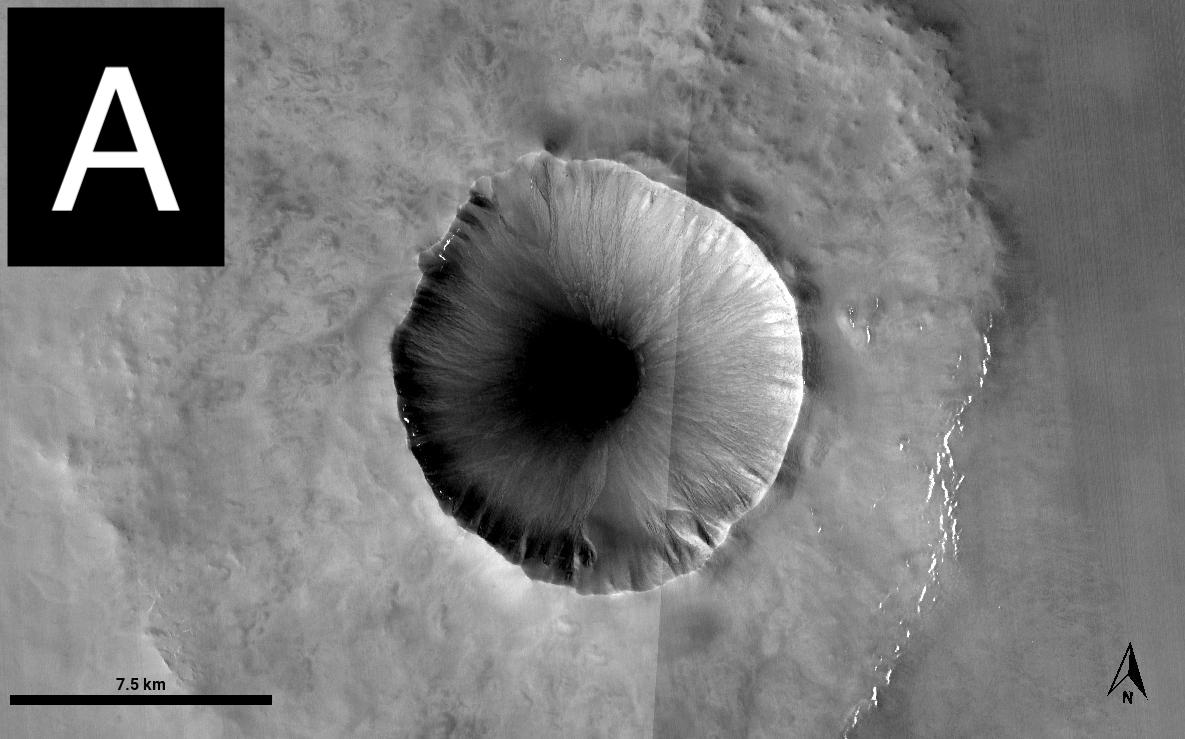}
    \includegraphics[width=0.19\textwidth]{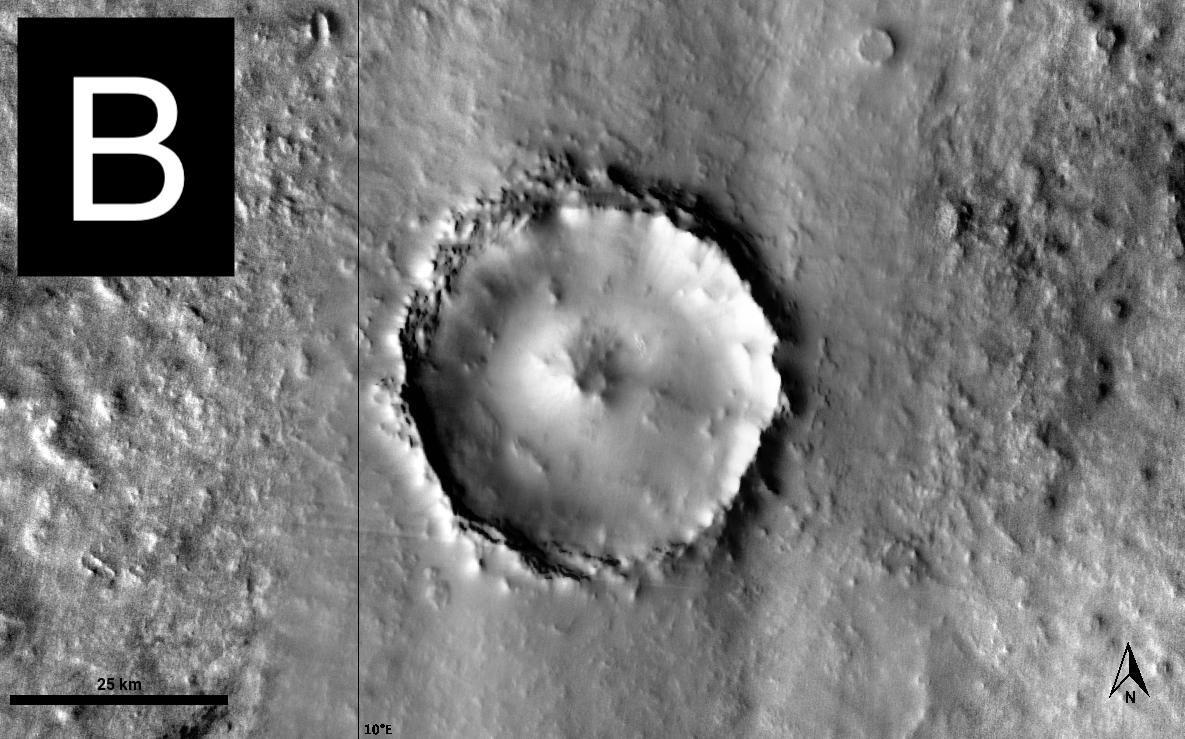}
    \includegraphics[width=0.19\textwidth]{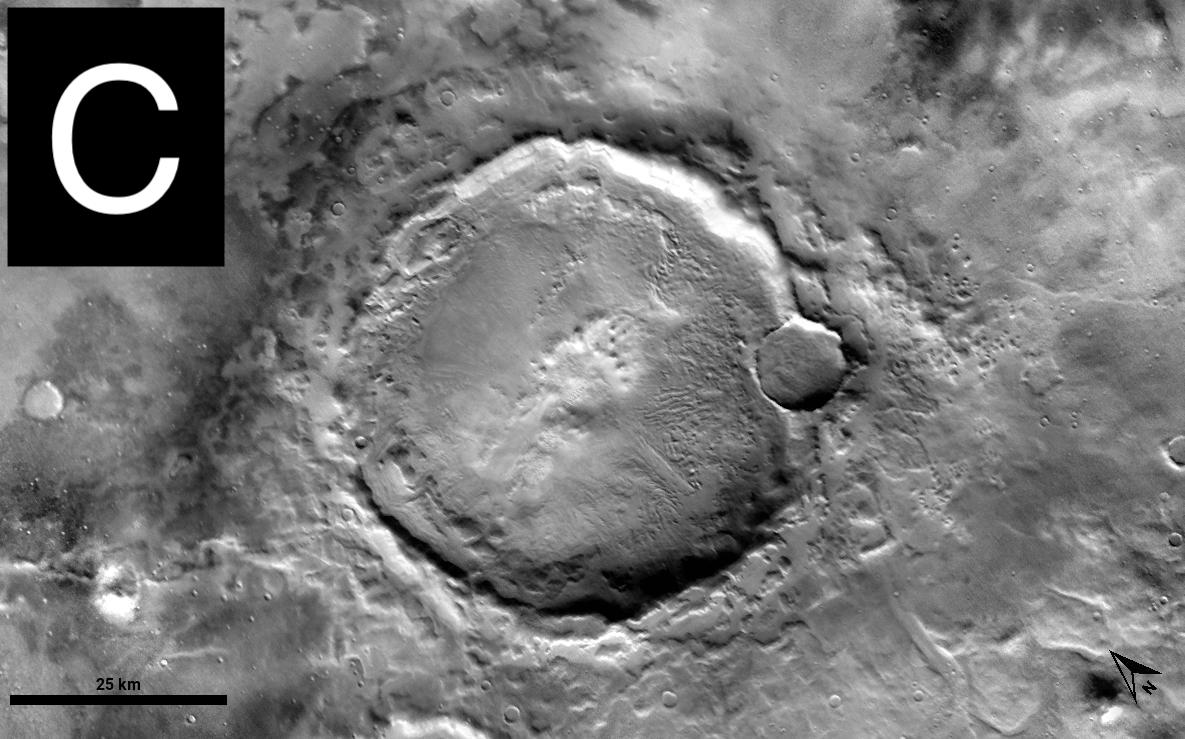}
    \includegraphics[width=0.19\textwidth]{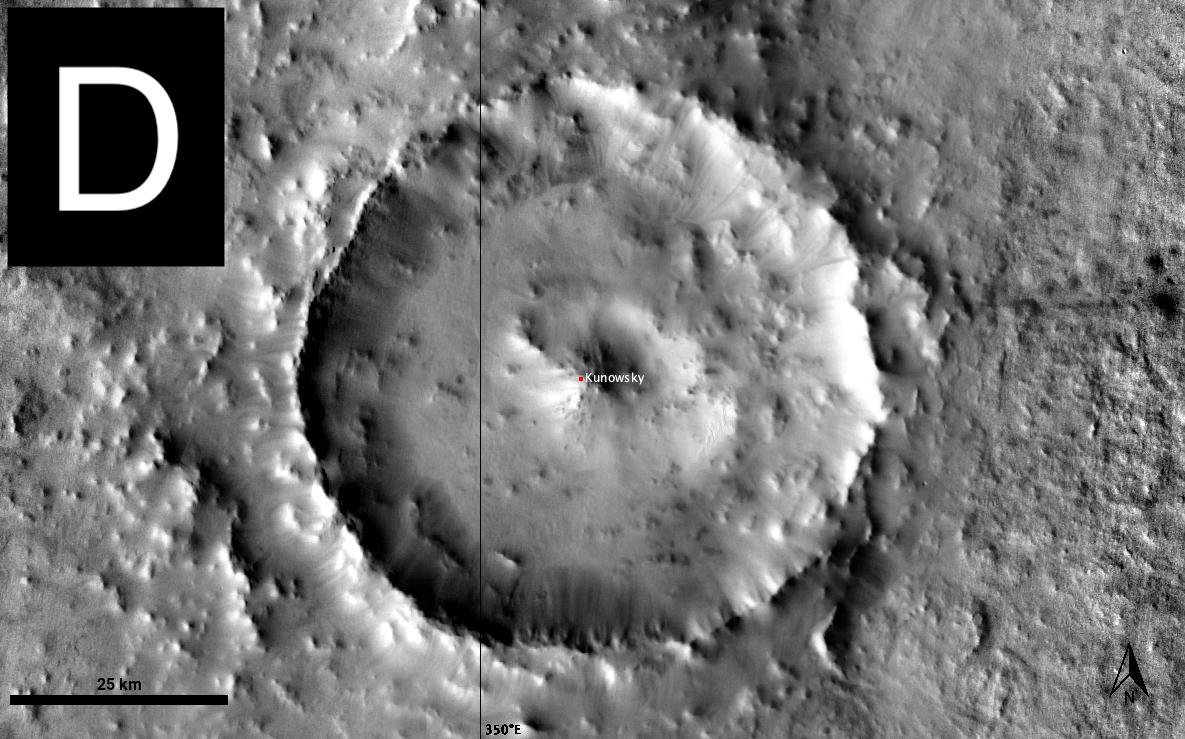} \\
    \includegraphics[width=0.19\textwidth]{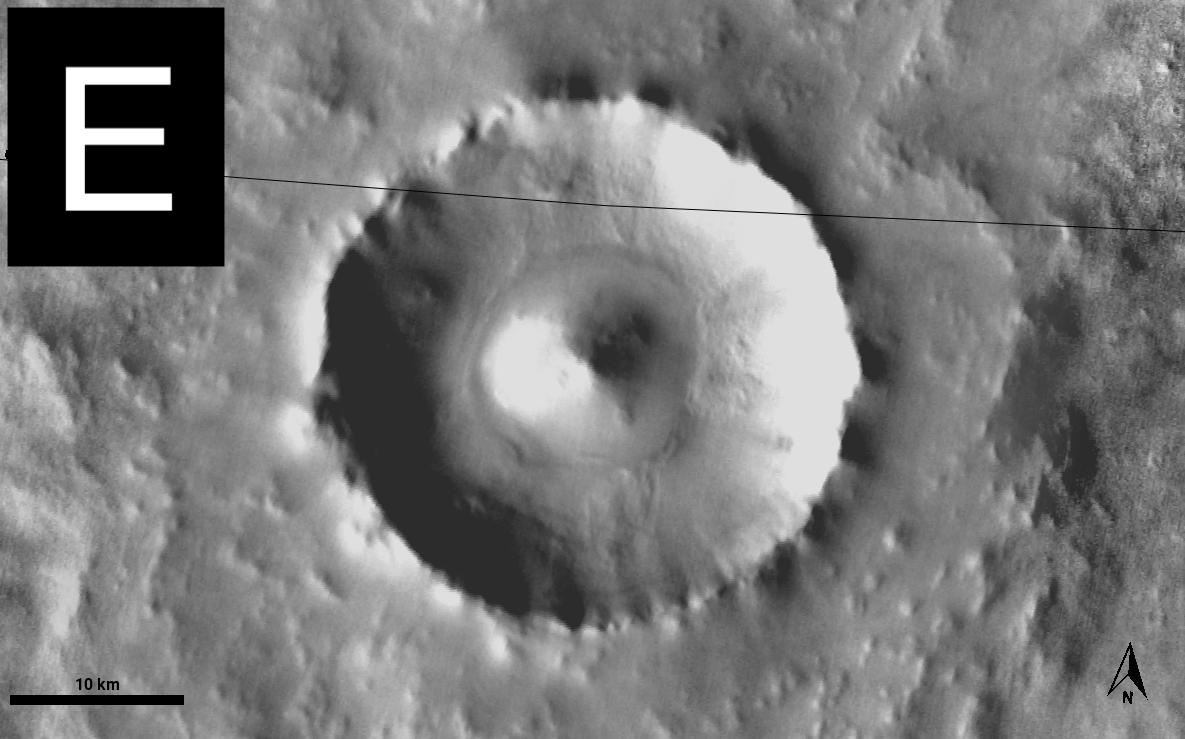}
    \includegraphics[width=0.19\textwidth]{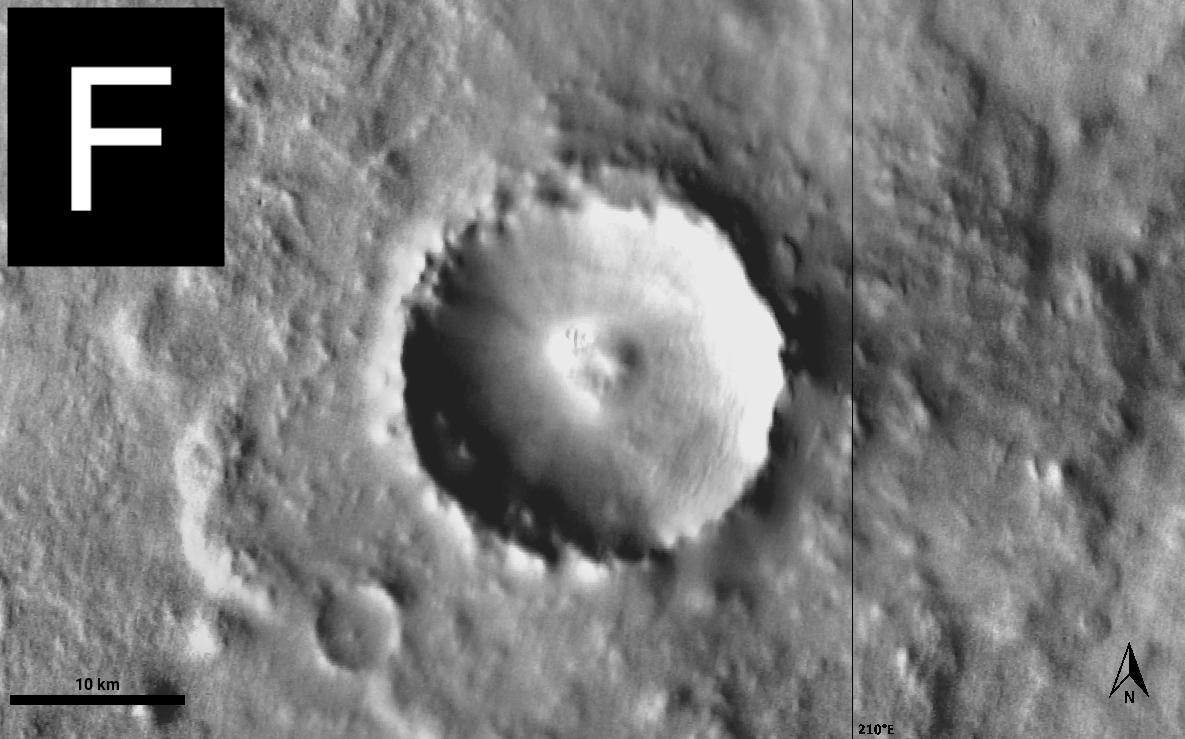}
    \includegraphics[width=0.19\textwidth]{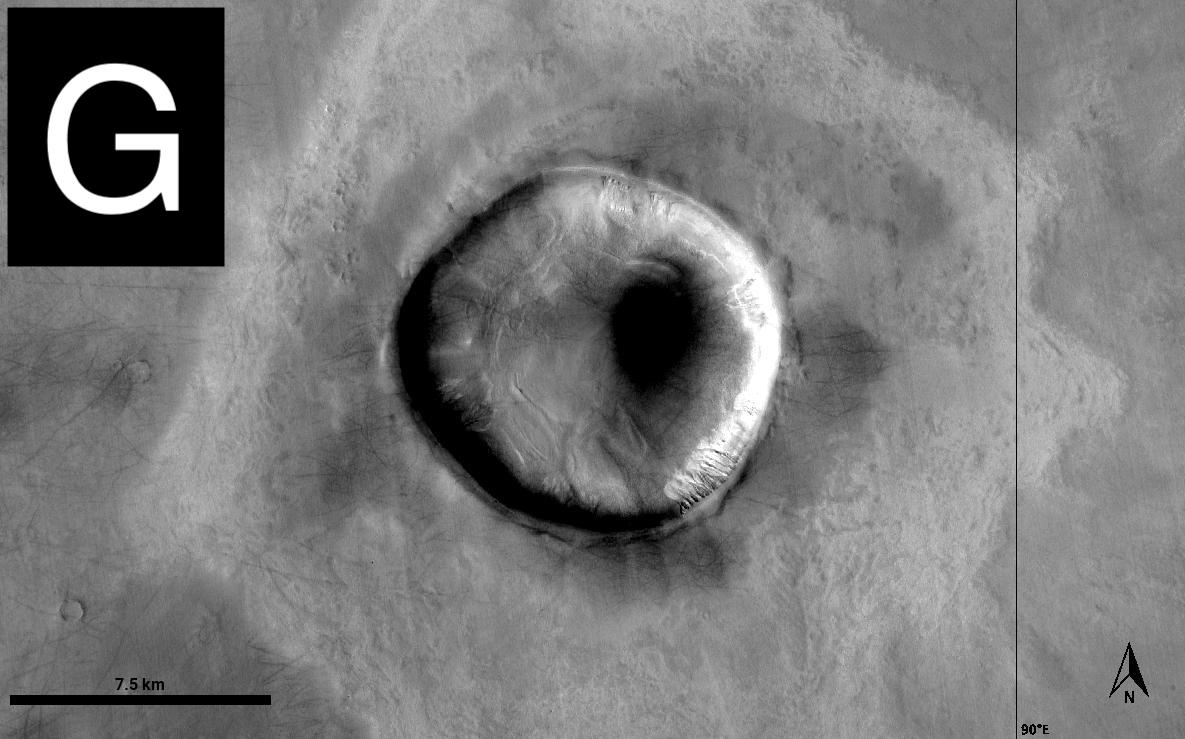}
    \caption{Previously studied northern mid-latitude frost sites used for training: %
    A:~\latlon{64.550}{315.907}, %
    B:~\latlon{58.236}{89.607}, %
    C:~\latlon{63.738}{11.035}, %
    D:~\latlon{42.572}{67.332}, %
    E:~\latlon{56.847}{350.401}, %
    F:~\latlon{59.839}{135.999}, %
    G:~\latlon{64.829}{209.406}.}
    \label{fig:nml-sites}
\end{figure*}

The scientific community is interested in better understanding the Martian global frost cycle, its effects on surface evolution, and its role in the larger Martian climate system. The extent of the contiguous seasonal frost cap and broad presence of frost has been mapped using low-resolution (\unit[0.1--6]{km/pixel}) thermal instruments~\cite{piqueux16diurnal}, while frost within specific small areas (including areas with patchier frost) has been investigated using medium-resolution (\unit[18]{m/pixel}) spectral instruments~\cite{pommerol11crismfrost} and high-resolution (\unit[25--50]{cm/pixel}) visible instruments~\cite{hansen10sublimation}. Due to the large volumes of data, extending the latter type of focused site studies globally, so as to combine that view with the global, low-resolution results, requires an automated approach. In this paper, we focus on training and evaluating a \ac{CNN} model for frost detection within observations acquired using the \ac{HiRISE} instrument~\cite{mcewen:hirise07}, which provides visible band, high-resolution surface images.

\begin{figure}
    \centering
    \includegraphics[width=0.2\textwidth,trim=0 0 0 256px, clip]{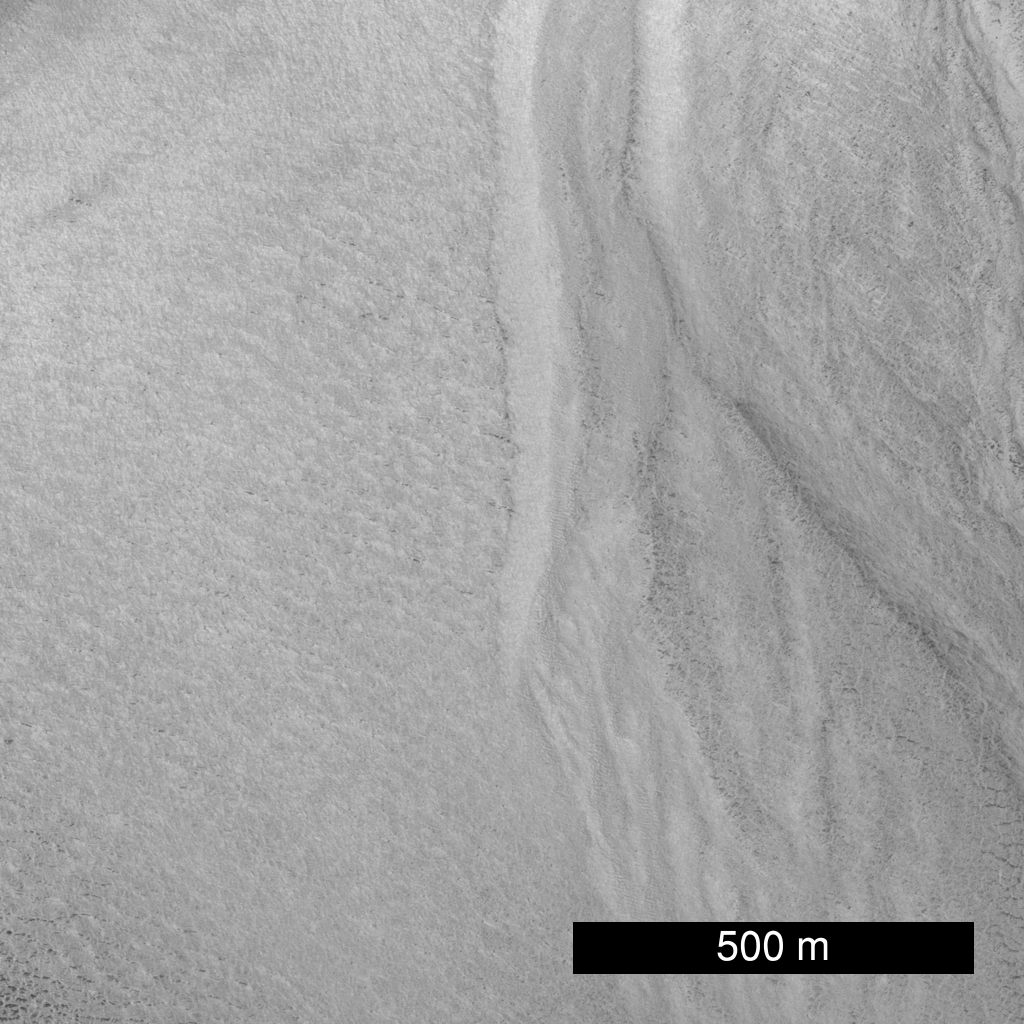}
    \includegraphics[width=0.2\textwidth,trim=0 0 0 256px, clip]{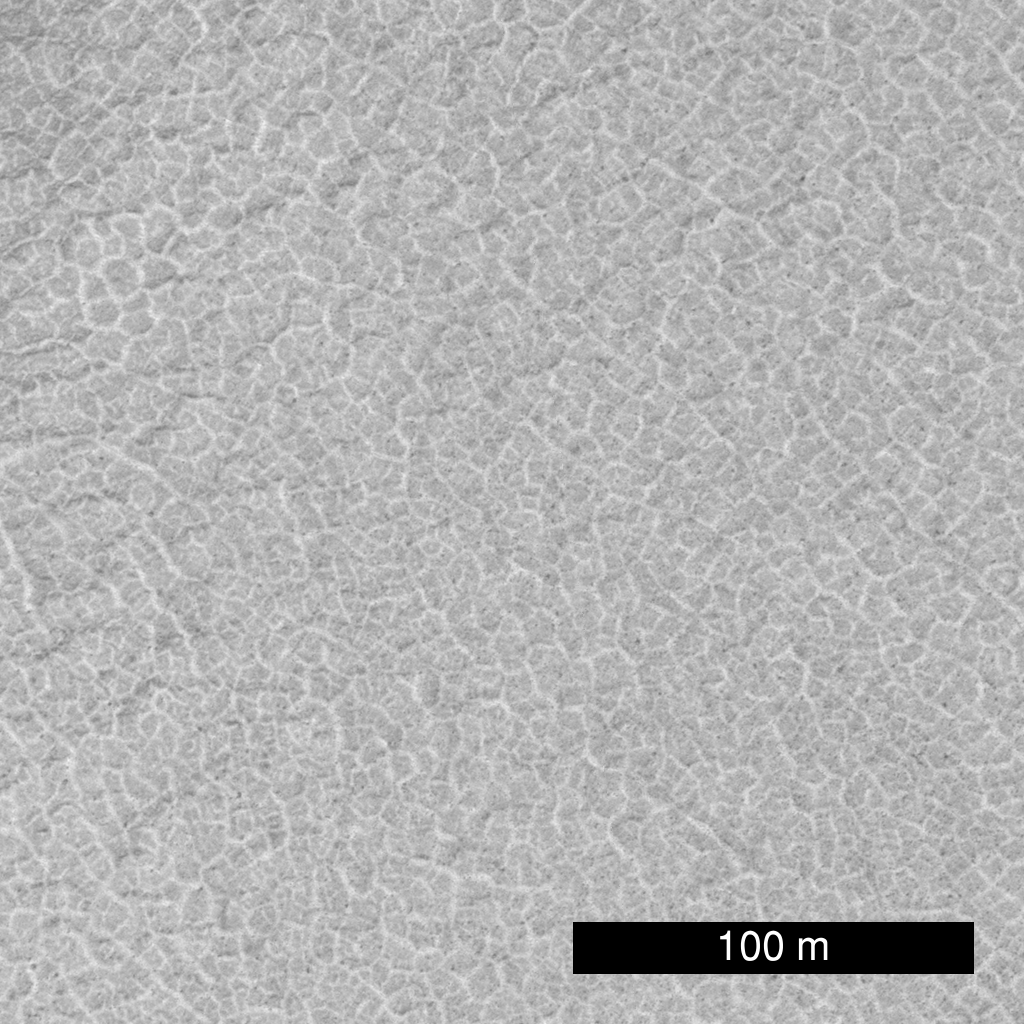}
    \includegraphics[width=0.2\textwidth,trim=0 0 0 256px, clip]{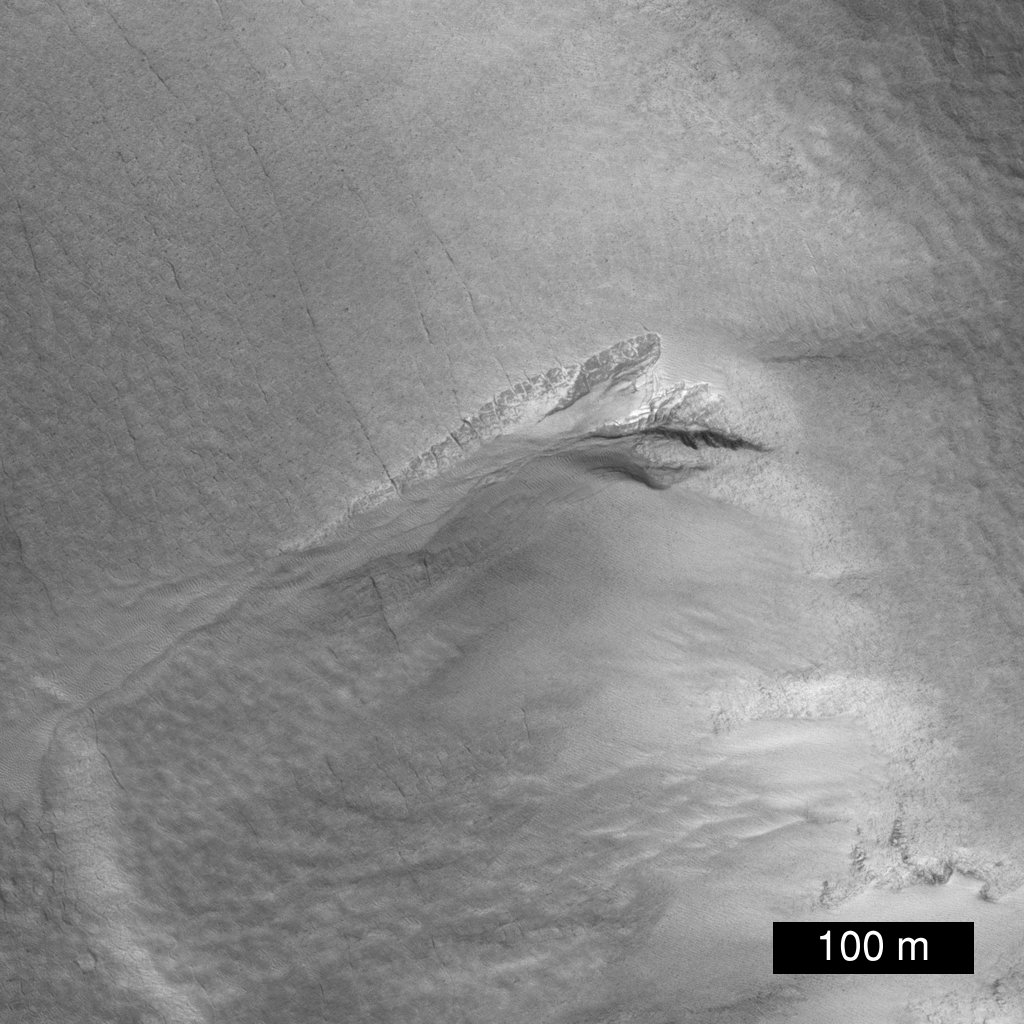}
    \includegraphics[width=0.2\textwidth,trim=0 0 0 256px, clip]{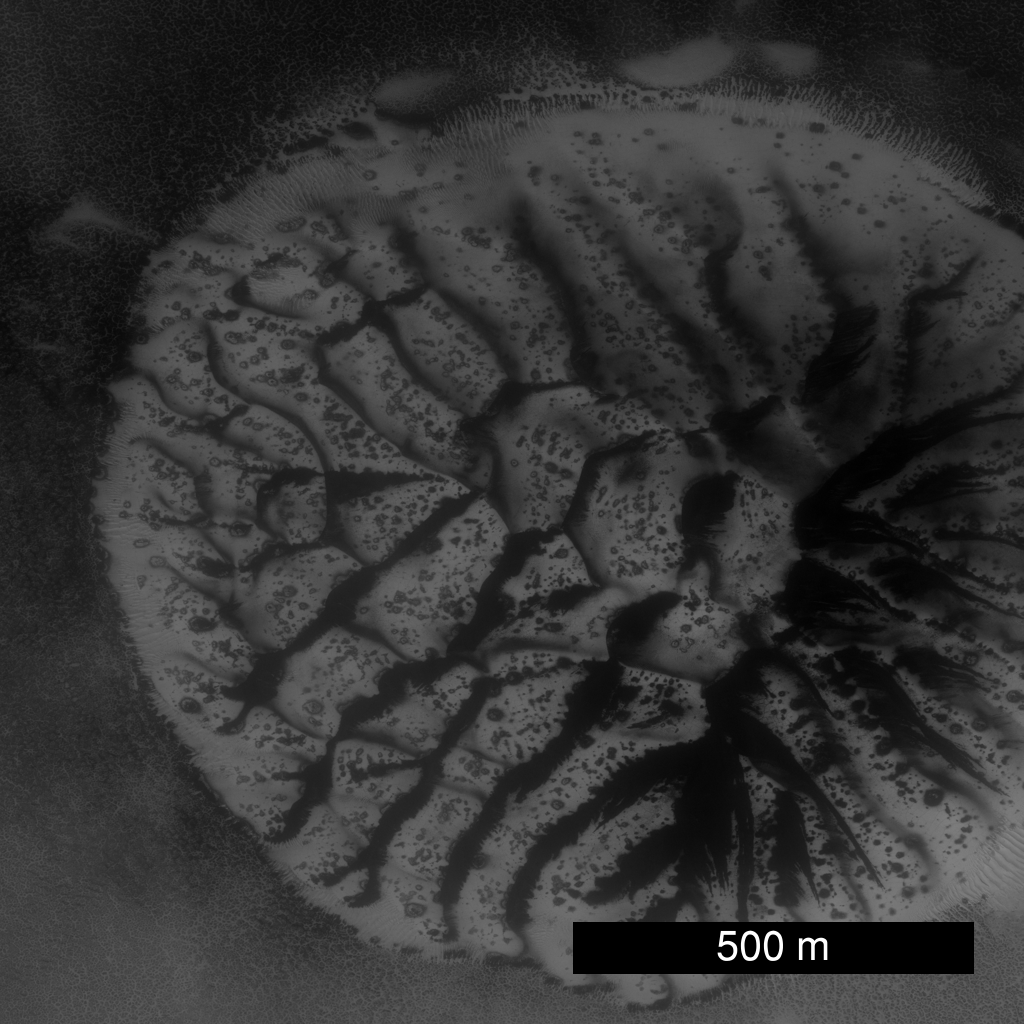}
    \caption{Visible indications of frost, including uniform albedo (top left), polygonal features (top right), halos (bottom left), and defrosting marks on dunes (bottom right).}
    \label{fig:frost-indicators}
\end{figure}

In order to train a \ac{CNN} frost detection model, we used \ac{HiRISE} observations collected for a previous frost study in the northern mid-latitude region of Mars~\cite{widmer2019frost}. The 7 sites we focused on (\autoref{fig:nml-sites}) are impact craters containing dark-colored basalt dune fields on which frost is visually apparent in the winter. The presence of frost on dunes indicates that regional conditions are favorable for frost formation, which helps to disambiguate whether frost may be present on nearby terrains. This aspect of the manual frost detection methodology highlights a key challenge for traditional machine learning models: confident frost detection often requires the use of larger scale contextual information not available to \ac{CNN} models using only local image information. To address this challenge, we focused on detecting \emph{visible indications} of frost, which include a uniform bright albedo, polygonal features, halos, and defrosting marks (see \autoref{fig:frost-indicators}) but excludes frost that is only detectable at other wavelengths~\cite{lange22frost}. This mirrors as closely as possible the manual frost detection problem, excluding the final step of assimilating information across scales and imaging modalities.

Our dataset consists of repeated observations of the same locations, which introduces another challenge when training models; we must account for the fact that the same locations are observed at different times throughout the seasonal cycle. In addition, because the data are highly clustered into discrete sites, it is necessary to account for these correlations during the validation process to prevent ``data leakage'' across the training, validation, and testing sets~\cite{kaufman12leakage,wadoux2021cv}. Below, we describe a novel spatial partitioning to address this challenge. One benefit of the overlapping images is that observations during summer months can be used to provide frost-free (negative) training examples, whereas observations during winter months provide candidate frost (positive) examples. 

\section{Methodology}

In this section, we provide an overview of our methodology from data generation to evaluation, with a focus on aspects specific to our dataset and problem domain.

\subsection{Creating a Machine Learning Ready Dataset}

Starting from \ac{HiRISE} observations of the northern mid-latitude sites shown in \autoref{fig:nml-sites}, we produced a set of labeled image tiles for the ``frost'' (positive) and ``background'' (negative) classes. Since determining frost composition is not straightforward using visible image data alone, we did not attempt to differentiate \ac{CO2} and \ac{H2O} frost. \ac{HiRISE} observations are typically around 10,000 pixels across, and of variable length depending on along-track exposure during ``push-broom'' imaging. We used the map-projected \ac{HiRISE} products available from the \ac{PDS}, which are between \unit[25--100]{cm/pixel} resolution~\cite{hirise}. Because entire observations are too large for most labeling tools, we break each observation into a ``subframe'' that is \unit[5,120]{pixels} along each dimension (except for partial subframes remaining near observation edges). Any subframe containing more than 75\% of pixels outside the valid map-projected data area were discarded. We randomly selected 15 subframes containing frost identified from previous studies and 15 without frost from each of the 7 sites, for a total of 210 subframes. Only the 105 frosted subframes required more detailed labeling.

We used the Labelbox\footnote{\url{https://labelbox.com/}} platform to annotate polygonal boundaries around regions with visible evidence of frost. For each polygon, we collected additional information from the labeler including the applicable visible indicators as well as geologic context, which is either ``dunes,'' ``gullies,'' ``crater rim/wall,'' or ``other.'' Here, the geologic context categories are mutually exclusive, so labelers could only pick one geologic context per frost polygon. We used the geologic context information to investigate terrain-dependent bias in classifier performance. To document the labeling process, we performed an iterative series of labeling sessions with both data science and science domain experts. Domain expert labeling guidance and clarifications at each iteration were captured in a labeling guide, included with the publicly available dataset\footnote{\url{http://doi.org/10.5281/zenodo.6561241}}. A total of 6 subframes, detailed in the released dataset, were excluded due to contamination with excess instrument noise or cloud cover. Each subframe was labeled by three different annotators.

Finally, we split each subframe into \unit[$299\times 299$]{pixel} tiles to generate a labeled dataset for training and evaluation. For each tile, the class label was determined through majority vote by comparing the set of overlapping polygons across the three annotators. If the number of polygons overlapping a tile is fewer than the required majority threshold, it was discarded to avoid ambiguous examples. The tiling process also discards any frost tiles that contain more than 10\% black pixels, which would indicate that they fall partially outside the valid map-projected image data.

\subsection{Spatial Validation}

\begin{figure}
    \centering
    \includegraphics[width=0.85\columnwidth]{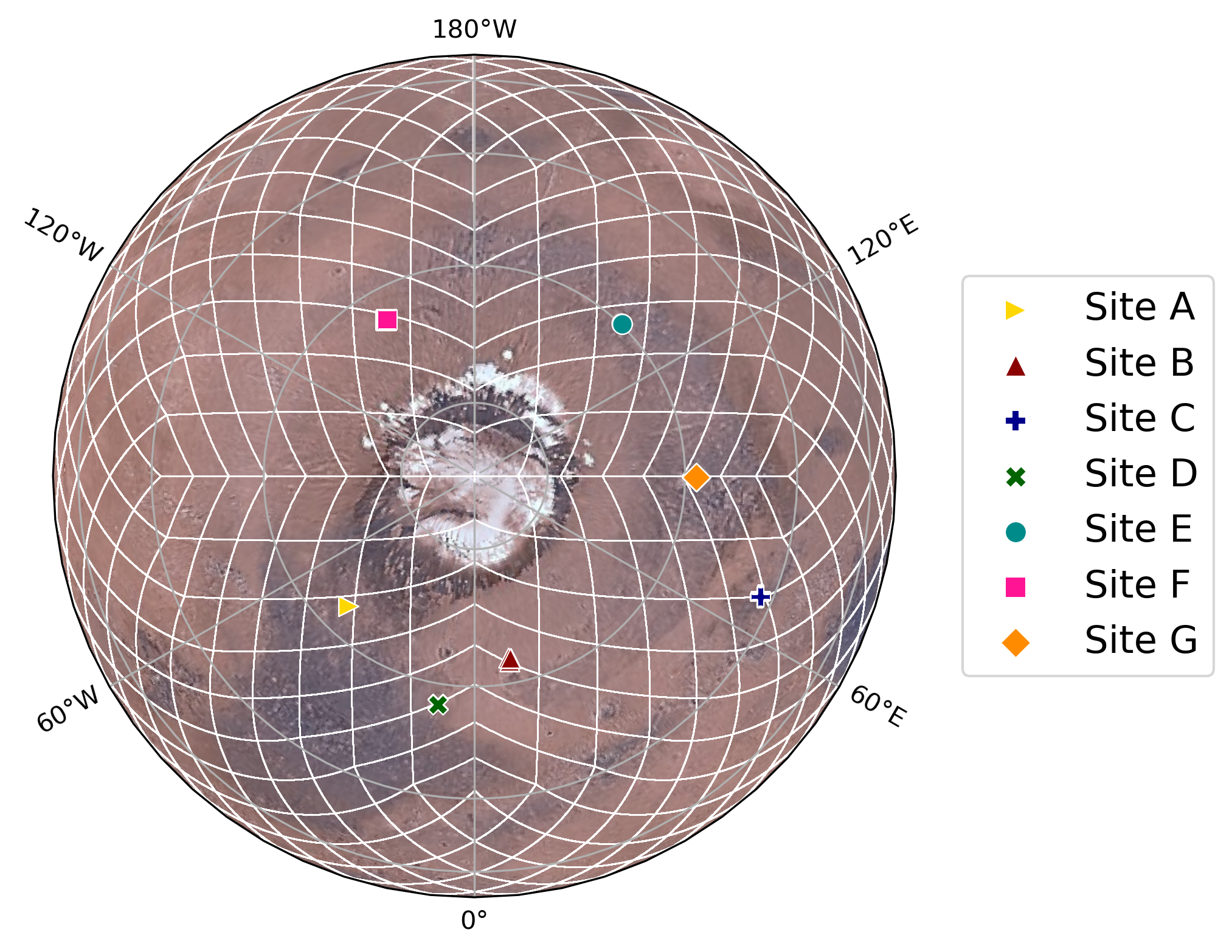}
    \caption{Site locations within the \acs{HEALPix} partition (white grid) of the surface (North pole orthographic projection).}
    \label{fig:healpix-cv}
\end{figure}

We used a standard train, validation, and test split methodology to evaluate the classifier both during and after training. Because the tiles derived from \ac{HiRISE} observations are spatially clustered, it is important to split the data in such a way that does not mix tiles from the same repeatedly imaged region across train, validation, and test folds. Here, we present a novel application of \ac{HEALPix} to spatially divide the globe into equal-area regions for model validation~\cite{gorski2005healpix}. The advantages of \ac{HEALPix} over other partitioning approaches is that it can be parameterized to flexibly subdivide the surface to arbitrary granularity ($N_{\mathrm{side}} = 8$ for this application, meaning that each base-resolution pixel is divided into 8 along each side). The equal-area nature of the pixelization also ensures that no regions are disproportionately represented. The pixelization used for our study, along with the location of the northern mid-latitude sites, is shown in \autoref{fig:healpix-cv}.

\subsection{Model Training and Evaluation}

We fine-tuned the InceptionV3 model~\cite{szegedy2015inception} for frost detection by adjusting the weights in the final fully-connected layer using TensorFlow~\cite{tensorflow2015whitepaper}. The learning rate was fixed to $10^{-3}$, and the batch size was set to 1. Training was performed for 100 epochs using the Adam optimizer~\cite{kingma2014adam} and cross-entropy loss, and the model with the best validation set accuracy was selected. We used classification accuracy to evaluate overall model performance on the training, validation, and test sets. To understand how geologic context affects detection for the positive frost class, we evaluated performance on the frost tiles using recall separately for each context.

\begin{table}
    \centering
    \begin{tabular}{|c|cccc|}
        \hline
        Context & Other  & Crater Rim/Wall & Gully & Dune  \\
        \hline
        Train   & 83.1\% & 10.0\% &  2.6\% &  4.3\% \\
        Test    & 98.3\% &  ---   &   ---  &  1.7\% \\
        \hline
    \end{tabular}
    \caption{Distribution of Geologic Context in Train/Test Sets}
    \label{tab:context-shift}
\end{table}

\begin{figure}
    \centering
    \includegraphics[width=0.45\textwidth]{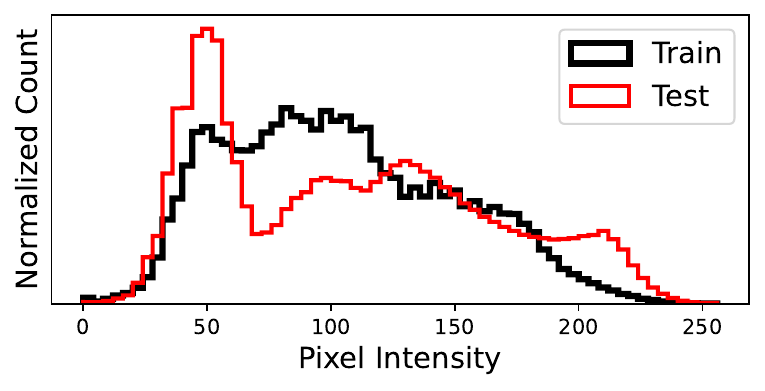}
    \caption{Comparison of pixel intensity distributions across all train and test tile pixels.}
    \label{fig:tile-hist}
\end{figure}

\begin{figure*}
    \centering
    \hfill\includegraphics[width=0.45\textwidth]{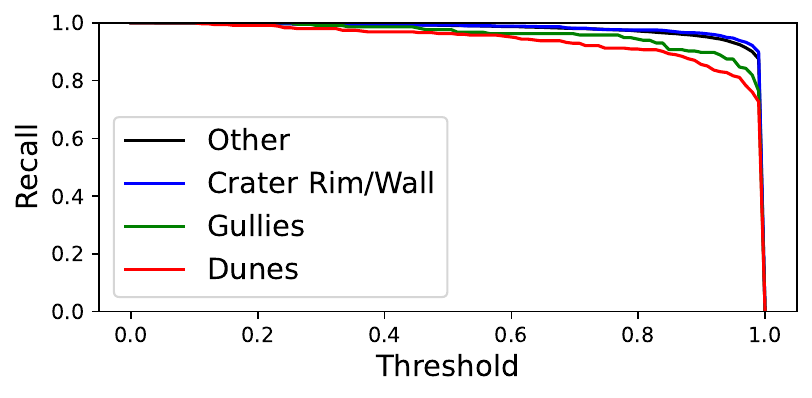}\hfill
    \includegraphics[width=0.45\textwidth]{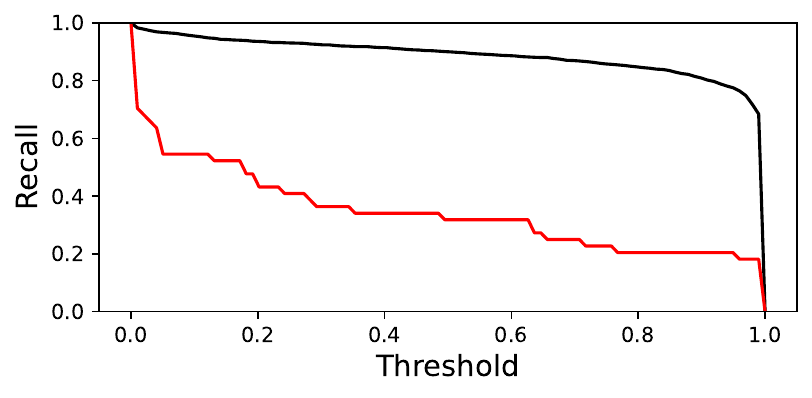}\hfill~
    \caption{Frost detection recall scores as a function classification threshold for train (left) and test (right) sets.}
    \label{fig:context-bias}
\end{figure*}
\section{Results}

The full labeled dataset contains 23,767 tiles, nearly balanced with 12,657
frost tiles and 11,110 background tiles (53.3\% frost tiles). During tile
generation, 6,471 potential frost tiles were excluded due to label ambiguity. Due to the \ac{HEALPix}-based spatial partitioning, there is some unevenness in splitting the data, so the training, validation, and test sets comprise 65.3\%, 14.3\%, and 20.4\% of the data set, respectively. Total labeling time across all three annotators was \unit[11.5]{hours}, which corresponds to \unit[1.7]{seconds} of labeling effort per tile produced.
The validation set accuracy was maximized on the $97^{th}$ training epoch with a value of 99.4\%. The selected model also has a training set accuracy of 99.4\% and a test set accuracy of 92.0\%. We hypothesize that the drop in accuracy on the test set relative to the validation set is in part due to inter-site variation in overall image tile characteristics.

\autoref{tab:context-shift} shows the distribution of geologic contexts (determined by majority vote across annotations) in both the train and test sets. The splitting of data by spatial partitioning induces a significant shift in this distribution in which two of the contexts (Crater Rim/Wall and Gully) are not represented in the test set. This suggests a relatively large degree of variability in terrain types covered by observations at each site. \autoref{fig:tile-hist} shows the overall differences in pixel intensity distributions across the train and test sets. While these distributions are similar, they do show some degree of overall covariate shift in addition to the shift in geologic context.

Focusing on context-dependent performance, \autoref{fig:context-bias} shows classifier recall on the train and test sets for each individual geologic context. Recall is plotted as a function of classification threshold. Even within the training set, there is reduced recall for some contexts, particularly gullies and dunes. Within the test set, there is a significant drop in recall for dunes relative to other contexts.

\begin{figure}[ht]
    \centering
    \includegraphics[width=0.45\textwidth]{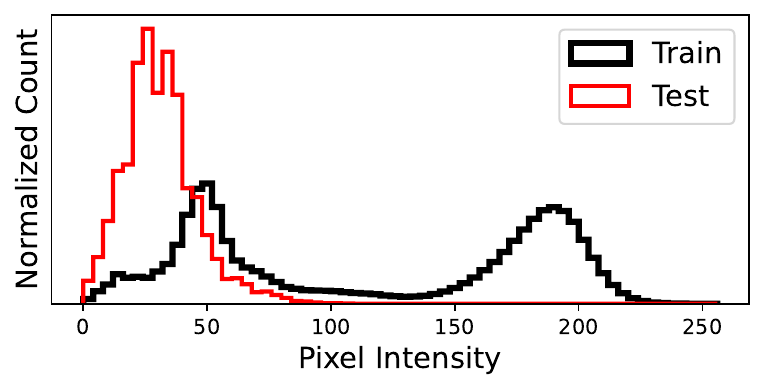} \\
    \begin{tabular}{c@{\hskip 5pt}c@{\hskip 15pt}c@{\hskip 15pt}c@{\hskip 5pt}c}
        \multicolumn{2}{c}{\Large{\textbf{Train Dunes}}} & & \multicolumn{2}{c}{\Large{\textbf{Test Dunes}}} \\
        \includegraphics[width=0.09\textwidth]{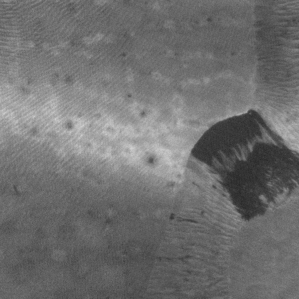} &
        \includegraphics[width=0.09\textwidth]{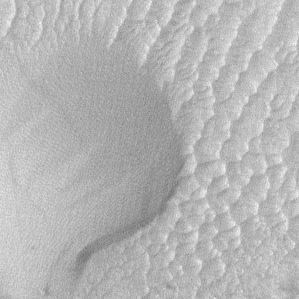} & &
        
        \includegraphics[width=0.09\textwidth]{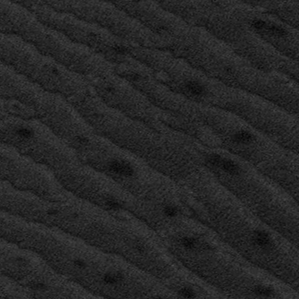} &
        \includegraphics[width=0.09\textwidth]{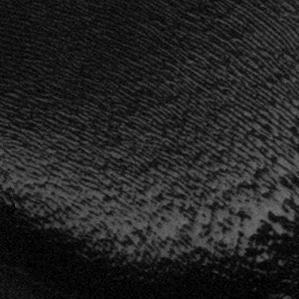} \\
        
        \includegraphics[width=0.09\textwidth]{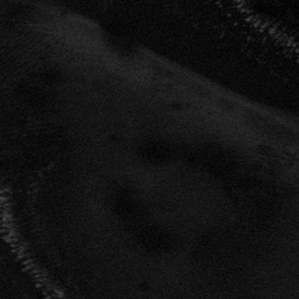} &
        \includegraphics[width=0.09\textwidth]{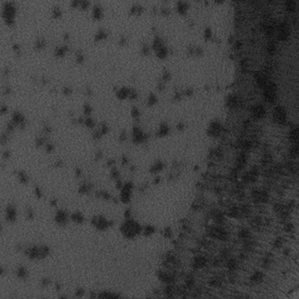} & &
        
        \includegraphics[width=0.09\textwidth]{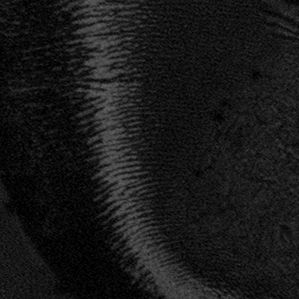} &
        \includegraphics[width=0.09\textwidth]{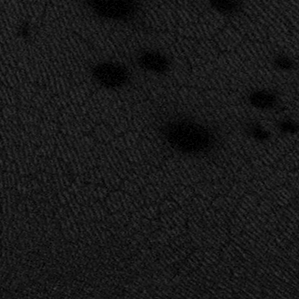} \\
        
    \end{tabular}
    \caption{(Top) Difference in pixel intensities across dune images within train and test sets. (Bottom) Representative examples of dune tiles in train and test sets.}
    \label{fig:dune-hist}
\end{figure}

To better understand the drop in frost recall scores on dunes, we visually inspect the differences in these tiles across the train and test sets, shown in \autoref{fig:dune-hist}. Overall, we see that the distribution of pixel intensities is multi-modal in the training set, with some overall bright and dark observations, whereas in the test set, only one mode is represented with darker pixels overall. The lower half of \autoref{fig:dune-hist} shows some representative examples from each set and shows a difference in the diversity of frost appearance across the two sets. These overall differences in brightness may be due to some combination of exposure and illumination across sites.

\section{Conclusion and Future Work}

In this work, we explore the use of ML models to automate the detection of surface frost in high-resolution Martian images. Specifically, we propose a new application of \ac{HEALPix} for partitioning spatial data to mitigate artificially inflated estimates of generalization performance across geologically varying sites on the surface of Mars. We explore the biases in model performance induced by the variability of observed geologic and observational characteristics across sites. In order to quantify this bias, collecting information about geologic context during labeling was an essential component of building a machine-learning-ready dataset for this domain.

We found that geologic context bias is present and significant for this model's performance on the test set, specifically for dune fields often found in northern mid-latitude craters. Interestingly, for human annotators, dunes often provide strong evidence of frost due to the striking visual appearance of defrosting marks which expose dark basalt sand beneath light-color frost. However, there is also a large degree of diversity in frost appearance on this underrepresented terrain type, both inherently and due to differing illumination and observational conditions, perhaps making the concept challenging for the classification model to learn.

To improve model performance and generalizability in future work, we propose (1) expanding the training set to include more diverse examples of underrepresented terrain types, (2) expanding the number of sites used to improve representation of all terrain types in the validation and test sets, and (3) performing contrast- and brightness-based augmentation to promote generalization under varying observational conditions. We expect these improvements will permit the training of models better suited for full-planet frost detection, thereby facilitating the creation of global frost maps.

\begin{acks}
\jplack{}
\end{acks}

\bibliographystyle{ACM-Reference-Format}
\bibliography{references}

\end{document}